\documentclass[10pt,twocolumn,letterpaper]{article}

\usepackage{cvpr}              % To produce the CAMERA-READY version

\usepackage{booktabs}
\usepackage{multirow}
\usepackage{makecell}
\usepackage{float}
\usepackage{tablefootnote}
\usepackage{footmisc}

\usepackage[linesnumbered,ruled,vlined]{algorithm2e}
\usepackage{algpseudocode}
\renewcommand{\algorithmicrequire}{\textbf{Input:}} 
\renewcommand{\algorithmicensure}{\textbf{Output:}} 

\SetCommentSty{mycommfont}

\newcommand{\pzero}{\phantom{0}}

\usepackage[dvipsnames]{xcolor}
\definecolor{cvprblue}{rgb}{0.21,0.49,0.74}
\usepackage[pagebackref,breaklinks,colorlinks,citecolor=cvprblue]{hyperref}

\def\paperID{14995}
\def\confName{CVPR}
\def\confYear{2024}

\title{Low-rank Attention Side-Tuning for Parameter-Efficient Fine-Tuning}

\author{Ningyuan Tang, Minghao Fu, Ke Zhu, Jianxin Wu\footnotemark[1]\\
National Key Laboratory for Novel Software Technology\\
School of Artificial Intelligence, Nanjing University, China\\
{\tt\small tangny@lamda.nju.edu.cn, fumh@lamda.nju.edu.cn, zhuk@lamda.nju.edu.cn, wujx2001@gmail.com}
}

\begin{document}
\maketitle

%将脚注符号设置为fnsymbol类型，即特殊符号表示
\renewcommand{\thefootnote}{\fnsymbol{footnote}} 
\footnotetext[1]{J. Wu is the corresponding author.}

\begin{abstract}
In finetuning a large pretrained model to downstream tasks, parameter-efficient fine-tuning (PEFT) methods can effectively finetune pretrained models with few trainable parameters, but suffer from high GPU memory consumption and slow training speed. Because learnable parameters from these methods are entangled with the pretrained model, gradients related to the frozen pretrained model's parameters have to be computed and stored during finetuning. We propose Low-rank Attention Side-Tuning (LAST), which disentangles the trainable module from the pretrained model by freezing not only parameters but also outputs of the pretrained network. LAST trains a side-network composed of only low-rank self-attention modules. By viewing the pretrained model as a frozen feature extractor, the side-network takes intermediate output from the pretrained model and focus on learning task-specific knowledge. We also show that LAST can be highly parallel across multiple optimization objectives, making it very efficient in downstream task adaptation, for example, in finding optimal hyperparameters. LAST outperforms previous state-of-the-art methods on VTAB-1K and other visual adaptation tasks with roughly only 30\% of GPU memory footprint and 60\% of training time compared to existing PEFT methods, but achieves significantly higher accuracy. 
\end{abstract}

\section{Introduction}

Finetuning large pretrained models on downstream tasks has become a widely-used paradigm in both natural language processing and computer vision \cite{raffel2020exploring,radford2021learning,he2022masked}. Thanks to knowledge or representation learned from the pretraining, finetuned models show good generalization ability on various downstream tasks. Compared to training a model from scratch, finetuning on a large pretrained model can both converge faster and achieve better performance.

\begin{figure}
    \centering
    \includegraphics[width=1.\linewidth]{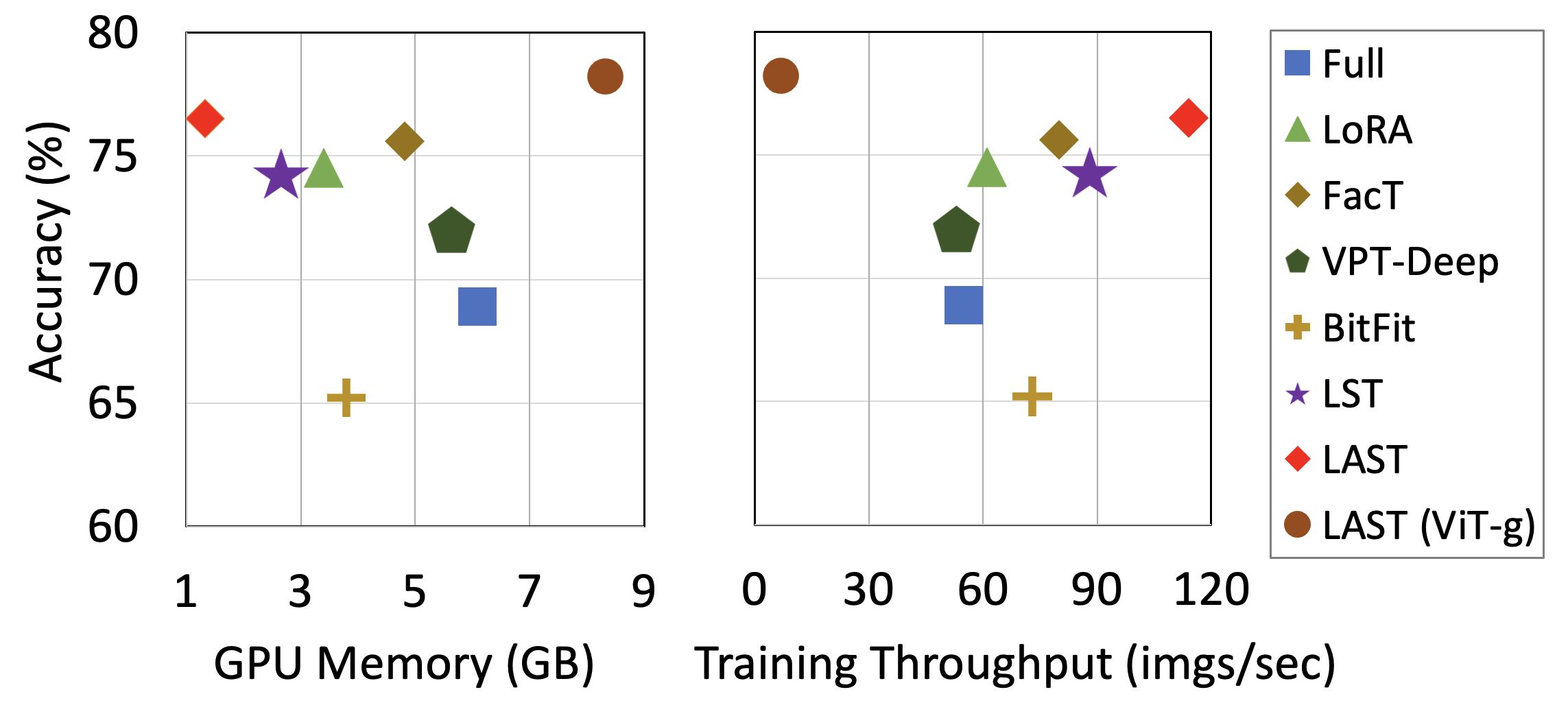}
    \caption{GPU memory footprint, accuracy, and training speed of different PEFT methods on the VTAB-1K~\cite{zhai2019large} benchmark. GPU memory and training throughput are tested with batch size 32 on one NVIDIA TITAN-Xp GPU. The proposed LAST method outperforms other methods with significantly lower GPU memory usage and higher training speed. The pretrained model is ViT-B. With its advantages, our LAST can even finetune the \emph{huge} ViT-g model~\cite{oquab2023dinov2} on \emph{single} NVIDIA TITAN Xp GPU (which has only 12GB memory and was released in year 2017)!}
    \label{fig:abstract}
\end{figure}

Parameter-efficient fine-tuning (PEFT) methods~\cite{houlsby2019parameter,li2021prefix,hu2021lora,he2023parameter,yin2023one,sung2022vl} aim to finetune large pretrained models with only a small number of trainable parameters, either by freezing the entire pretrained model but add few learnable parameters, or by freezing most but a tiny subset of the pretrained model's parameters. The benefit of PEFT is obvious: the storage cost of finetuned models is significantly reduced, and fewer trainable parameters makes them easier to train. Meanwhile, models are less likely to overfit on downstream tasks. On many adaptation tasks, PEFT methods~\cite{jia2022visual,hu2021lora,lian2022scaling} can perform close to or even better than full model finetuning with less than 1\% trainable parameter.

In spite of their successes in achieving high accuracy with few trainable parameters, drawbacks of PEFT methods are also clear: the finetuning process still endures \emph{large GPU memory consumption} and \emph{slow training speed}. The reason is that the computations of the frozen part and the learnable parameters are \emph{entangled} together, such that gradients beyond the trainable parameters (\ie, with respect to the big pretrained model) need to be computed and stored~\cite{sung2022lst}. As shown in Figure~\ref{fig:abstract}, LoRA~\cite{hu2021lora} uses nearly half of the GPU memory and almost the same training time as those of the full model finetuning!

On the other hand, side-tuning methods, which fully separates the computations involving the pretrained model from a ``side-network'', does not need to cache any gradient for pretrained model~\cite{zhang2020side,sung2022lst}. That is, in side-tuning, the pretrained network can be treated as a standalone feature extractor. Hence side-tuning naturally uses fewer GPU memory and trains faster than other PEFT methods. But, these methods are not as competitive as non-side-tuning PEFT methods in terms of both accuracy and parameter-efficiency (\ie, requires more learnable parameters).

In this paper, we are in defense of side-tuning for parameter-efficient fine-tuning. We propose such an approach that is both more parameter-efficient and achieves finetuning accuracy even higher than state-of-the-art non-side-tuning methods. Our method keeps enjoying the benefits in small GPU memory and fast training. As Figure~\ref{fig:abstract} shows, with \emph{single outdated TITAN Xp} GPU, we can finetune the huge ViT-g model with batch size 32.

Our key finding is that for PEFT, if we use Transformer blocks as learnable modules in side-tuning, we do \emph{not} need the large feed-forward network, while the self-attention module can have a \emph{extremely low rank} (\eg, 4 or 8). Hence, the proposed method is called LAST (Low-rank Attention Side-Tuning). Our contributions can be summarized as follows:
\begin{itemize}
	\item We propose LAST, a side-tuning framework for PEFT, which exhibits virtues of both side-tuning and non-side-tuning PEFT methods, including small GPU memory footprint, fast training/finetuning, parameter-efficient, and achieves state-of-the-art PEFT accuracy.
    \item The key in the proposed LAST framework is the LSA (low-rank self-attention) module, which for the first time utilizes low-dimensional self-attention and throws away the huge feed-forward network for side-tuning. We also correct the LSA bias to make it work properly for PEFT. 
    \item Furthermore, the proposed method enables us to finetune multiple LAST models with different sets of hyperparameters in parallel, which greatly facilitates hyperparameter searching in PEFT.
\end{itemize}

\section{Related Works}

In this section, we briefly review related prior works on parameter-efficient fine-tuning and side-tuning.

\subsection{Parameter Efficient Fine-Tuning (PEFT)}

PEFT is widely used in both NLP and computer vision, which aims to adapt a pretrained model to a downstream task by only modifying or adding very few trainable parameters. 

Adapter~\cite{houlsby2019parameter} introduces a lightweight finetune method by inserting learnable MLP layers to Transformer blocks after each attention~\cite{vaswani2017attention} and FFN layer. AdaptFormer~\cite{chen2022adaptformer} inserts MLP layers to FFN in a parallel form. VPT~\cite{jia2022visual} does not insert parameters to the pretrained network, but adds task specific learnable prompt tokens to finetune on downstream tasks. There are two variants: VPT-Shallow and VPT-Deep. VPT-Shallow only inserts prompt tokens before the first block, while VPT-Deep inserts prompt tokens before each Transformer block. LoRA~\cite{hu2021lora} proposes to finetune low-rank decomposition matrices of a dense layer instead of finetuning dense layers directly. Since the low-rank decomposition matrices can be merged into dense layers, LoRA does not change the model architecture and will not introduce additional inference latency. 
% SSF~\cite{lian2022scaling} transforms intermediate features with a scaling factor $\gamma$ and a shifting factor $\beta$ after each MHSA~\cite{vaswani2017attention}, FFN and LayerNorm~\cite{ba2016layer} layer. 
Existing PEFT methods, however, still consumes large chunks of GPU memory and is slow in the finetuning process, even though they only contain few learnable parameters.

\subsection{Side-Tuning}

Side-tuning is proposed in~\cite{zhang2020side} to alleviate catastrophic forgetting in transfer learning. \cite{zhang2020side} keeps the backbone frozen and initializes a new side-network with two strategies: if the side-network has the same form as the backbone, the backbone parameters are directly copied to the side-network, otherwise the side-network is initialized by knowledge distillation~\cite{hinton2015distilling} from the backbone. Both networks take images as input and only the side-network is trainable. Output of the side-tuning is the additive outcome of the backbone output and the side-network output.

Ladder Side-Tuning (LST)~\cite{sung2022lst} takes inspiration from side-tuning, but they reduce the size of side block to achieve memory-efficient finetuning. LST adopts the T5 block~\cite{raffel2020exploring} as its side-network block. Similar to side-tuning, LST initializes the side block by knowledge distillation. LST takes block-wise intermediate features from the frozen backbone and linearly project the features to match the dimensionality of the side-network. The projected intermediate features are added to the side-network to provide additional input. LST achieves memory-efficiency by freezing the backbone and reducing the size of the side-network. However, its parameter efficiency is relatively low (\ie, requires more learnable parameters), and more importantly, its PEFT accuracy is lower than state-of-the-art non-side-tuning methods.

\section{In Defense of Side-Tuning for PEFT}

Before introducing the proposed LAST method for PEFT, we start by arguing that side-tuning is a desired framework for parameter-efficient fine-tuning with more details.

Side-tuning refers to a method to adapt a pretrained base model to some specific downstream tasks~\cite{zhang2020side}. Given a pretrained network $B$ and an input $x$, it produces $B(x)$ as the representation for $x$. However, because $B$ is not designed for the downstream task, a task-specific network (often much smaller than $B$) $S$ will be trained, and the final representation will then be $\alpha B(x)+(1-\alpha)S(x)$, where $\alpha$ is a hyperparameter that linearly combines pretrained and task-specific information.

In side-tuning, the pretrained model $B$ and task-specific one $S$ are \emph{independent} of each other, meaning the training of $S$ does not involve  the network parameters in $B$, \emph{nor its computation graph}. In other words, $B$ can be viewed as a standalone feature extractor. This particular choice leads to many benefits~\cite{zhang2020side}, with a notable one being that side-tuning does not suffer from catastrophic forgetting. When there are multiple downstream tasks, this property is essential.

However, by requiring $B$ and $S$ to be independent, this specific form of side-tuning also highly restricts the representation power of $S(x)$, because it has to learn the task-specific side-network $S$ on its own. The LST method~\cite{sung2022lst} proposes a ladder structure, where the pretrained network $B$ will occasionally send its activations (features) as inputs to $S$, such that the training process of $S$ can benefit from the pretrained representation. 

\begin{figure*}
    \centering
    \includegraphics[width=0.9\linewidth]{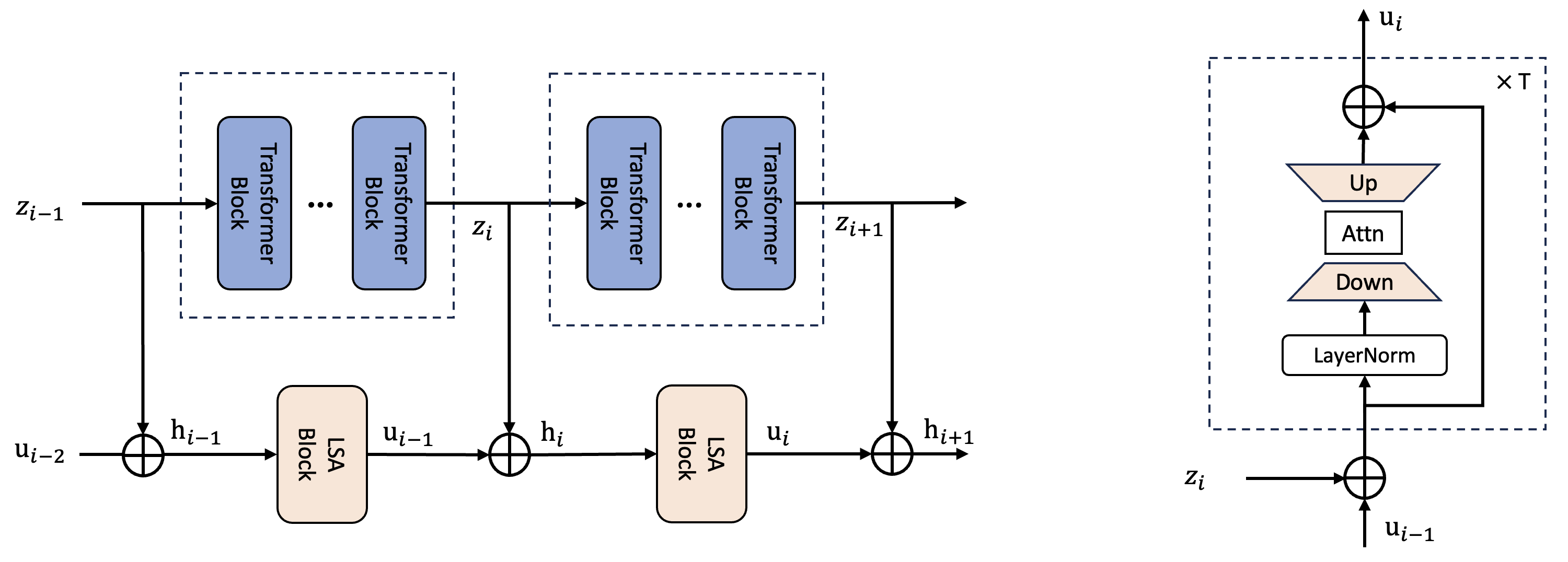}
    \caption{Overall architecture of the proposed LAST method. In the left, for a group of $g$ Transformer blocks, we insert an LSA block in the side-network, whose details are in the right part of this figure. Note that there is \emph{no} arrow from the LSA modules (bottom branch) back to the pretrained network (top branch), hence the pretrained network can be viewed as a \emph{standalone feature extractor}.}
    \label{fig:structure}
\end{figure*}

As illustrated in Figure~\ref{fig:structure} (left), the arrows from the top branch ($B$) to the bottom branch ($S$) helps the learning of $S$. It is worth emphasizing that \emph{there is no arrow from $S$ pointing back to $B$}, hence, the forward computation of $B(x)$ is \emph{not} affected by $S$. Furthermore, there is \emph{no} backward gradient computation in the pretrained model ($B$) branch. As will be discussed later, this fact helps save a lot of GPU memory consumption during training $S$. Instead of $S(x)$, we use $S(x,B(x))$ to denote the side-network's representation.

Main-stream PEFT methods often add or modify a small subset of parameters to change $B$ into $B'$, but freeze all the pretrained model's parameters or its non-modified subset. By viewing these modified or added parameters or modules (such as the LoRA modules~\cite{hu2021lora}) as a side-branch $S$, they affect the computation of later network blocks in $B'$, because they amount to \emph{add arrows from the bottom task-specific branch ($S$) to the top branch ($B$)}. In other words, the final representation will be $B'(x)=f(B(x),S(x))$, where $f$ is a \emph{highly non-linear} function that mixes the computations and representations of $B$ and $S$. Hence, they may suffer from catastrophic forgetting if not properly handled. And, this strategy requires backward gradient computation in learning $B$/$B'$.

Hence, ideally we are in defense of the side-tuning strategy, where the final representation can be \emph{linearly separated} into two parts: the standalone pretrained features $B(x)$, and the task-specific part $S(x,B(x))$.

But, side-tuning (or LST) has not been the main-stream method for PEFT, probably due to the fact that this line of methods lead to \emph{lower accuracy} and is \emph{less parameter-efficient} in downstream tasks. Next, we propose our side-tuning framework, which is not only parameter-efficient but also achieves significantly higher accuracy than existing methods in vision PEFT tasks.

\section{LAST: Low-rank Attention Side-Tuning}
\label{ssec:LAST}

Our key finding in achieving this goal is that in order to adapt a pretrained model $B$ to a downstream task, the side-network $S$ \emph{only needs very low-rank self-attention} if we adopt attention blocks in $S$, and that the feed-forward network (FFN) is \emph{not} required. Accordingly, we name our method LAST, or Low-rank Attention Side-Tuning. The key component of LAST is a new low-rank self-attention (LSA) module (Figure~\ref{fig:structure}, right).

As illustrated in Figure~\ref{fig:structure}, LAST groups $g$ blocks in the pretrained model $B$ as a unit, and inserts an LSA block in the task specific side-network $S$ along with each such unit. The LSA block takes two inputs: one from the previous LSA block, and the other from the base network $B$.

Formally, suppose the pretrained model $B$ has $N$ blocks. We first define a gap factor $g$ where $N$ is divisible by $g$. The base pretrained model $B$ is divided into $m=N/g$ groups, and hence there will be $m$ LSA blocks in $S$, correspondingly. Note that $B$ is completely frozen, and we adopt Vision Transformers as the pretrained model. We denote tokens after $B$'s patch embedding layer as $z_0$, and the tokens after (\ie, the output of) the $i$-th group in $B$ as $z_i$.

Corresponding to these groups in $B$, LSA blocks are inserted into $S$ after $z_0,z_1,\dots,z_m$, respectively, as shown in Figure~\ref{fig:structure}. One LSA block consists of $T$ LSA modules, with each containing layer normalization~\cite{ba2016layer}, our low-rank attention, and residual connections~\cite{he2016residual,wang2019learning,baevski2018adaptive}.

The computation in the $i$-th LSA block, $F_i(\cdot)$, can be defined as:
\begin{equation}
    F_i(x)=f_i^T(f_i^{T-1}(\cdots f_i^1(x))) \,,
\end{equation}
where $f_i^j(\cdot)$ ($1 \le j \le T$) is an LSA module (proposed in Section~\ref{ssec:lsa}). $F_i$ takes two input: the output from both the previous LSA block's output $u_{i-1}$, and the intermediate features $z_i$ from $B$:
\begin{align}
    h_i &= u_{i-1} + z_{i} \,,\\
    u_i &= F_i(h_i) \,.
\end{align}

\subsection{Low-rank Self-Attention}
\label{ssec:lsa}

The proposed LSA module is based on standard ViTs~\cite{dosovitskiy2020image}, and we first briefly introduce the MHSA (multi-head self-attention) module in ViT. 

For input tokens $X$, MHSA first projects $X$ into the triplet of query, key, and value ($Q$, $K$, and $V$) by projections using learnable weights $W_Q$, $W_K$, and $W_V$, respectively:
\begin{equation}
    (Q,K,V)=(XW_Q,XW_K,XW_V) \,.
\end{equation}
Note that all these tokens are $d$ dimensional. These tokens are evenly split into $n$ heads, e.g., $Q$ is split into $Q_1, Q_2, \dots, Q_n$, with $d/n$ dimensions in each token in $Q_i$. Then, self-attention is carried out in the $i$-th head as:
\begin{equation}
 \mathrm{head}_i = \mathrm{Attention}(Q_i,K_i,V_i)=\mathrm{softmax}\left(\dfrac{Q_iK_i^\top}{\sqrt{d/n}}\right)V_i\,,
\end{equation}
and these outputs are concatenated and projected again by learnable weights $W_O$:
\begin{equation}
    \mathrm{MHSA}(Q,K,V)=\mathrm{Concat}({\rm head}_1,\cdots,{\rm head}_{n})W_O \,.
\end{equation}
Note that in ViTs, one Transformer block contains not only MHSA, but also a very large feed-forward network (FFN).

However, when adapting a pretrained model to downstream tasks, it is often not optimal to finetune the entire network $B$, because the target domain is much simpler than the pretraining domain and the downstream training set is usually very small. By simply adding very few parameters~\cite{chen2022adaptformer} to $B$ or modifying a tiny subset of parameters in $B$~\cite{zaken2021bitfit}, parameter-efficient fine-tuning methods can not only achieve parameter-efficiency, but also generally obtain higher accuracy than full model finetuning.

In other words, the difference between the pretraining and the downstream task's representation seems pretty small. Inspired by these observations, we reasonably hypothesize that we \emph{can throw away large FFN (}\cf. \emph{Appendix for details), and low-rank self-attention is sufficient} if we use Transformer-like architecture in the side-network. Thus, we propose our low-rank self-attention (LSA) module, which have \emph{very small dimensionality in the self-attention module}.

For input tokens, we first project the tokens (after layer normalization) into the $Q$, $K$, $V$ triplet, but with very low dimensionality $r$ (\ie, $r \ll d$, \eg, $r=16$ when $d=768$), which is illustrated as `Down' in Figure~\ref{fig:structure}. Then, the MHSA operation is carried out on the low-dimensional projected tokens. Finally, the tokens are projected back to $d$ dimensions (illustrated as `Up' in Figure~\ref{fig:structure}). 

The LSA module is described in Algorithm~\ref{alg:lsa}. Note that since we have the `Up' projection, $W_O$ is not needed. In practice, we find that the dimensionality of each attention head ($r/n$) can be as small as 4 or even 2 while achieving high accuracy in PEFT tasks.

\begin{algorithm}
  \caption{The proposed LSA module}
  \label{alg:lsa}
	\algorithmicrequire{ Input tokens $X$, down projection weights $A_Q$, $A_K$, $A_V$, up projection weights $B$}\\
	\Begin{
	$X'\leftarrow{\rm LN}(X)$ \tcp*{layer norm} 
	$Q,K,V \leftarrow X'A_Q,X'A_K,X'A_V$ \tcp*{`Down'} 
	$X'\leftarrow{\rm MHSA}(Q,K,V)$ \tcp*{attention} 
	$X'\leftarrow X'B$  \tcp*{`Up'} 
	\algorithmicensure{ $X+X'$ \tcp*{residual connection} }
	}
\end{algorithm}

It is worth emphasizing that to the best of our knowledge, we are the \emph{first} to find the utility of self-attention with very low dimensionality (\ie, $r \ll d$) and its surprising effectiveness for downstream vision tasks.

\subsection{Correcting the LSA Bias}
\label{ssec:subtract}

Up to now, the final representation in our framework is $u_m$, \ie, output of the final LSA block. As aforementioned, we want $u_m$ to be the sum of two components: one from the pretrained network $B$ ($z_m$ in our notation), and another task-specific one from the side-network $S$. 

Now consider any LSA module $f_i^1$ (\ie, the first LSA module in the $i$-th LSA block). Its input $X=u_{i-1} + z_i$. According to Algorithm~\ref{alg:lsa}, its output is $X+X'=z_i + u_{i-1} + X'$. Note that both $u_{i-1}$ and $X'$ are task-specific, because their computations involve the side-network. Now it is clear that $f_i^1(X)$ can be decomposed as the sum of $z_i$ (pretrained) and a task-dependent term. By induction, the output of $F_i$ (composition of $f_i^1,f_i^2,\dots,f_i^T$) can be decomposed as the sum of $z_i$ (pretrained) and a task-dependent term, too.

Again, by induction, it is easy to prove that the final representation, $u_m$, can be decomposed as the sum of the following two terms:
\begin{enumerate}
	\item One pretrained term $z_0+z_1+\dots+z_{m}$;
	\item Another task-specific term from the side-network $S$.
\end{enumerate}

The discrepancy is obvious: We want the pretrained term to be $z_m$, but $\sum_{i=0}^m z_i$ is erroneously provided by LSA. This bias in LSA is caused by the residual connection, which is essential and cannot be removed. Hence, we correct this bias by defining the final representation of LAST as
\begin{equation}
	u_m - \sum_{i=0}^{m-1}z_i \,.
\end{equation}

One final technical note is on the initialization of the LSA module. For LoRA~\cite{hu2021lora}, which also have `Up' and `Down' projections, the `Down' projection is randomly initialized but the `Up' projection is zero initialized, in order to make the initial network (with additional LoRA modules) remain the same as the pretrained backbone. This initialization strategy facilitates finetuning the entire model (backbone + LoRA modules). However, in LAST, since the pretrained network does not take part in back-propagation, we randomly initialize both these projections.

\newcommand\term[1]{\textsc{Term}~\textsc{#1}}

\subsection{Advantages of LAST and Side-Tuning}

The proposed LAST method has advantages in various aspects when compared to existing PEFT methods: much smaller GPU memory footprint, higher PEFT accuracy, and faster training.

\subsubsection{GPU memory footprint and accuracy}
\label{ssec:memory}

When finetuning a pretrained model, three parts add up to the total memory usage: model parameters, input data, and computational graph cached for back propagation. GPU memory consumption of the first term depends on model size, second term on data size in a mini-batch, and the third term is proportion to the product of the previous two. Hence, the third term is often the dominant factor in memory consumption. Given a linear layer being the $i$-th layer with weights $W_i\in\mathbb{R}^{d\times d}$ and nonlinear activation function $\sigma_i$, \cite{sung2022lst} shows that the gradient for $W_{i}$ with respect to the loss $L$ is
\begin{equation}
    \dfrac{\partial L}{\partial W_i}=\underbrace{\dfrac{\partial L}{\partial \textbf{a}_{i+1}}}_{\term{a}}\cdot\underbrace{\dfrac{\partial \textbf{a}_{i+1}}{\partial \textbf{u}_i}}_{\term{b}}\cdot\underbrace{\dfrac{\partial \textbf{u}_i}{\partial W_i}}_{\term{c}} \,,
    \label{eq:back}
\end{equation}
where $\textbf{u}_i=W_i\textbf{a}_i$ and $\textbf{a}_{i+1}=\sigma_i(\textbf{u}_i)$. Similarly, this analysis can be extended to other layers like self-attention. Most parameter-efficient fine-tuning (PEFT) methods aim to reduce the number of trainable parameters, which indirectly decreases the total size of $\term{c}$ but cannot affect $\term{b}$ in Eq.~\ref{eq:back}. 
% For example, LoRA~\cite{hu2021lora} introduces a low rank matrix $\Delta W=BA$ to finetune $W\in\mathbb{R}^{d\times k}$, where $B\in\mathbb{R}^{d\times r}$, $A\in\mathbb{R}^{r\times k}$ and $r\ll \min(d,k)$. This reduces the size of $\term{c}$ from $d\cdot k$ to $r\cdot(d+k)$, but $\term{b}$ ($d \times d$ in size) remains unchanged.

As long as a PEFT method interweaves the computation of the pretrained network $B$ (\ie, when $B$ is no longer a standalone feature extractor) and the side-network $S$, intermediate activation maps of $B$ must be cached, which consumes huge chunks of GPU memory. On the other hand, side-tuning methods (including LST~\cite{sung2022lst} and our LAST) do not need to cache them, which greatly reduces GPU memory footprint and enables usage of \emph{much larger pretrained models} to increase PEFT accuracy.

Compared to LST~\cite{sung2022lst}, LAST is more parameter-efficient, whose number of learnable parameters is only less than 1/3 of that in LST, which will be verified by our experiments in Section~\ref{sec:exp}. More importantly, its finetuning accuracy is significantly higher than not only LST but state-of-the-art non-side-network methods (\eg,~\cite{lian2022scaling}), as will be shown by our experimental results, too. Note that non-side-network methods require much higher GPU footprint during finetuning.

\subsubsection{Faster training \& parallel training}
\label{ssec:parallel}

As aforementioned, the pretrained backbone $B$ does not need any gradient back-propagation in our LAST, which greatly accelerates the training (finetuning) process. Furthermore, since our self-attention is very low-rank, the side-network is a tiny one and the finetuning bottleneck is actually the forward-only computation of $B$. This makes LAST roughly 2 times faster in terms of finetuning speed when compared to other PEFT methods (\cf. Fig.~\ref{fig:abstract}). Note that although LoRA has added few trainable parameters, its finetuning speed is only slightly faster than a full finetuning.

Moreover, the architecture of LAST is highly efficient for parallel training. Because the pretrained network $B$ can be viewed as a standalone feature extractor, one can extract the tokens $z_0,z_1,\dots,z_m$ in one forward computation, after which the side-network $S$ can be trained without the pretrained network $B$. Hence, \emph{many} small different side-networks can be trained \emph{simultaneously}, with different weights, hyperparameters and optimization objectives.

A common and very useful scenario for this type of parallel training is when one wants to search the optimal set of hyperparameters by finetuning many different side-networks with different hyperparameters and potential network structures simultaneously. In our experiments, we employed this parallel training scheme to find the optimal set of hyperparameters for LAST, and highly accelerates not only the final model finetuning with the chosen set of hyperparameters, but also the entire learning process.

\begin{table*}
         \centering
         \setlength{\tabcolsep}{1.5pt}
         \small
         \begin{tabular}{l|ccccccc|cccc|cccccccc|ccc}
         \toprule[1.5pt]
			& \multicolumn{7}{c|}{\textbf{Natural}} & \multicolumn{4}{c|}{\textbf{Specialized}} &  \multicolumn{8}{c|}{\textbf{Structured}} \\ 
   % \midrule
			  & \rotatebox{90}{CIFAR-100} & \rotatebox{90}{Caltech101} & \rotatebox{90}{DTD} & \rotatebox{90}{Flowers102} & \rotatebox{90}{Pets} & \rotatebox{90}{SVHN}  & \rotatebox{90}{Sun397} & \rotatebox{90}{Patch Camelyon~} & \rotatebox{90}{EuroSAT}   & \rotatebox{90}{Resisc45}  & \rotatebox{90}{Retinopathy} & \rotatebox{90}{Clevr/count} & \rotatebox{90}{Clevr/distance}  & \rotatebox{90}{DMLab} & \rotatebox{90}{KITTI/distance~}  & \rotatebox{90}{dSprites/loc} & \rotatebox{90}{dSprites/ori}   & \rotatebox{90}{SmallNORB/azi~}  & \rotatebox{90}{SmallNORB/ele~} & \rotatebox{90}{GPU Mem (GB)} & \rotatebox{90}{Params (M)} & \rotatebox{90}{Mean Acc} \\
			\midrule
			Full finetuning & 68.9 & 87.7 & 64.3 & 97.2 & 86.9 & 87.4 & 38.8 & 79.7 & 95.7 & 84.2 & 73.9 & 56.3 & 58.6 & 41.7 & 65.5 & 57.5 & 46.7 & 25.7 & 29.1 & 6.09 & 85.8 & 68.9 \\
			Linear probing & 64.4 & 85.0 & 63.2 & 97.0 & 86.3 & 36.6 & 51.0 & 78.5 & 87.5 & 68.5 & 74.0 & 34.3 & 30.6 & 33.2 & 55.4 & 12.5 & 20.0 & \pzero9.6 & 19.2 & 0.57 & 0.0 & 57.6 \\
			\midrule
                BitFit~\cite{zaken2021bitfit} & 72.8 & 87.0 & 59.2 & 97.5 & 85.3 & 59.9 & 51.4 & 78.7 & 91.6 & 72.9 & 69.8 & 61.5 & 55.6 & 32.4 & 55.9 & 66.6 & 40.0 & 15.7 & 25.1 & 3.80 & 0.10 & 65.2 \\
                VPT~\cite{jia2022visual} & \textbf{78.8} & 90.8 & 65.8 & 98.0 & 88.3 & 78.1 & 49.6 & 81.8 & 96.1 & 83.4 & 68.4 & 68.5 & 60.0 & 46.5 & 72.8 & 73.6 & 47.9 & 32.9 & 37.8 & 5.63 & 0.56 & 72.0 \\
                COMPACTOR~\cite{karimi2021compacter} & 71.9 & 89.0 & 69.7 & 99.1 & 90.7 & 82.7 & \textbf{56.1} & 86.0 & 93.5 & 82.4 & 75.3 & 80.2 & 63.4 & 47.4 & 77.2 & 78.1 & \textbf{53.5} & 27.3 & 39.8 & 4.39 & \textbf{0.04} & 74.2 \\
                LST~\cite{sung2022lst} & 59.5 & 91.5 & 69.0 & 99.2 & 89.9 & 79.5 & 54.6 & \textbf{86.9} & 95.9 & 85.3 & 74.1 & 81.8 & 61.8 & \textbf{52.2} & 81.0 & 71.7 & 49.5 & \textbf{33.7} & 45.2 & 2.65 & 2.38 & 74.3 \\
                LoRA~\cite{hu2021lora} & 67.1 & 91.4 & 69.4 & 98.8 & 90.4 & 85.3 & 54.0 & 84.9 & 95.3 & 84.4 & 73.6 & \textbf{82.9} & \textbf{69.2} & 49.8 & 78.5 & 75.7 & 47.1 & 31.0 & 44.0 & 3.40 & 0.29 & 74.5 \\
                AdaptFormer~\cite{chen2022adaptformer} & 70.8 & 91.2 & 70.5 & 99.1 & \textbf{90.9} & 86.6 & 54.8 & 83.0 & 95.8 & 84.4 & \textbf{76.3} & 81.9 & 64.3 & 49.3 & 80.3 & 76.3 & 45.7 & 31.7 & 41.1 & 4.11 & 0.16 & 74.7 \\
                FacT~\cite{jie2023fact} & 70.6 & 90.6 & 70.8 & 99.1 & 90.7 & \textbf{88.6} & 54.1 & 84.8 & 96.2 & 84.5 & 75.7 & 82.6 & 68.2 & 49.8 & 80.7 & 80.8 & 47.4 & 33.2 & 43.0 & 4.81 & 0.07 & 75.6 \\
                % SSF \cite{lian2022scaling} & 69.0 & 92.6 & 75.1 & 99.4 & 91.8 & 90.2 & 52.9 & 87.4 & 95.9 & 87.4 & 75.5 & 75.9 & 62.3 & 53.3 & 80.6 & 77.3 & 54.9 & 29.5 & 37.9 & 5.59 & 0.21 & 75.7 \\
			\midrule
                % LAST-Lite \textbf{(ours)}\\
                LAST \textbf{(ours)} & 66.7 & \textbf{93.4} & \textbf{76.1} & \textbf{99.6} & 89.8 & 86.1 & 54.3 & 86.2 & \textbf{96.3} & \textbf{86.8} & 75.4 & 81.9 & 65.9 & 49.4 & \textbf{82.6} & \textbf{87.9} & 46.7 & 32.3 & \textbf{51.5} & \textbf{1.33} & 0.66 & \textbf{76.5} \\
			\bottomrule
		\end{tabular}
	
	\caption{Results on the VTAB-1K benchmark with ViT-B/16 pretrained on IN21k. ``Params'' denotes the number of trainable parameters. ``GPU Mem'' denotes the GPU Memory footprint when finetuning with batch size 32. ``Mean Acc'' is the group-wise average accuracy over three task groups.}
	\label{tab:cmp_vtab}
\end{table*}

\begin{table*}
    \centering
	\small
    \setlength{\tabcolsep}{3pt}
    \begin{tabular}{l|ccccc|c}
        \toprule
         Method & CUB-200 & NABirds & \makecell{Oxford\\Flowers} &\makecell{Stanford\\Dogs} & \makecell{Stanford\\Cars} & \makecell{Mean\\Acc} \\
        \midrule
            Full finetuning & 87.3 & 82.7 & 98.8 & 89.4 & 84.5 & 88.5 \\
            Linear probing & 85.3 & 75.9 & 97.9 & 86.2 & 51.3 & 79.3 \\
            Adapter~\cite{he2023parameter} & 87.1 & 84.3 & 98.5 & 89.8 & 68.6 & 85.7 \\
            VPT-Shallow~\cite{jia2022visual} & 86.7 & 78.8 & 98.4 & \textbf{90.7} & 68.7 & 84.6 \\ 
            VPT-Deep~\cite{jia2022visual} & \textbf{88.5} & 84.2 & 99.0 & 90.2 & 83.6 & 89.1 \\
        \midrule
            LAST \textbf{(ours)} & \textbf{88.5} & \textbf{84.4} & \textbf{99.7} & 86.0 & \textbf{88.9} &  \textbf{89.5} \\
        \bottomrule 
    \end{tabular}
    \caption{Results on fine-grained visual classification tasks with ViT-B/16 backbone pretrained on IN21k.}
    \label{tab:cmp_other}
\end{table*}

\begin{table*}
         \centering
         \setlength{\tabcolsep}{1.5pt}
         \small
         \begin{tabular}{l|ccc|ccc|ccc}
         \toprule[1.5pt]
			& \multicolumn{3}{c|}{\textbf{ViT-B (IN21K)}} & \multicolumn{3}{c|}{\textbf{ViT-L (IN21K)}} &  \multicolumn{3}{c}{\textbf{ViT-g (LVD-142M)}} \\ 
   % \midrule
			  & Params (M) & GPU Mem (G) & Mean Acc & Params (M) & GPU Mem (G) & Mean Acc & Params (M) & GPU Mem (G) & Mean Acc \\
			\midrule
			Full & 85.8 & 6.09 & 68.9 & 307 & 15.0 & 72.7 & 1167 & - & - \\
			Linear & \pzero0 & 0.57 & 57.6 & \pzero\pzero0 & \pzero1.47 & 65.0 & \pzero\pzero\pzero0 & 5.00 & 67.1 \\
                \midrule
                LST~\cite{sung2022lst} & 2.38 & 2.65 & 74.3 & 8.17 & \pzero5.16 & 74.3 & 30.26 & 20.41 & 77.3 \\
                VPT~\cite{jia2022visual} & 0.56 & 5.63 & 72.0 & 1.23 & \pzero9.34 & 74.5 & - & - & - \\
                LoRA~\cite{hu2021lora} & \textbf{0.29} & 3.40 & 74.5 & \textbf{0.77} & \pzero9.19 & 75.8 & - & - & - \\
                LAST \textbf{(ours)} & 0.66 & \textbf{1.33} & \textbf{76.5} & 1.63 & \textbf{\pzero3.35} & \textbf{76.9} & \phantom{0}\textbf{4.13} & \phantom{0}\textbf{8.63} & \textbf{78.2} \\
			\bottomrule
		\end{tabular}
	
	\caption{Results on VTAB-1K with backbones of different sizes. ViT-B and ViT-L are pretrained on ImageNet-21k~\cite{imagenet}, and ViT-g is pretrained on LVD-142M~\cite{oquab2023dinov2}. ``Params'' denotes the number of trainable parameters. ``GPU Mem'' denotes the GPU Memory footprint when finetuning with batch size 32. ``Mean Acc'' is the group-wise average accuracy over three task groups of VTAB-1K. All experiments are restricted to an NVIDIA GeForce RTX 3090 GPU, and methods out of memory are marked with a hyphen (`-') in the table.}
	\label{tab:cmp_arch}
\end{table*}

\section{Experiments} \label{sec:exp}

In this section, we evaluate the performance of LAST. We compare LAST with previous PEFT methods on various visual domains (VTAB-1K~\cite{zhai2019large} and various FGVC datasets). After that we verify the scalability of LAST on pretrained models of different sizes (\eg, ViT-B, ViT-L, ViT-g). Finally, ablation studies are conducted to further analyze the impact of different designs in LAST.

\subsection{Implementation Details}

We conducted most of our experiments on a ViT-B/16 pretrained with IN21k. In the main experiments, if not additionally mentioned, we set $g=2$, $T=2$ for ViT-B and ViT-L, and $g=4$, $T=2$ for ViT-g. For ViT-B and ViT-L, the hidden dimensionality of low-rank self-attention is set to $r=16$, with 4 heads ($n_{\rm head}=4$) and each head's dimensionality is $r_{\rm head}=4$. For ViT-g, we set $r=32$, $r_{\rm head}=4$ and $n_{\rm head}=8$. For a fair comparison, following the setting of~\cite{jia2022visual}, images were directly resized to $224\times224$ for VTAB-1K, and we performed a random resize crop with random horizontal flip for FGVC (fine-grained visual classification) datasets. For both VTAB-1K and FGVC datasets, we used the Adam optimizer~\cite{kingma2014adam} with batch size 32 and trained the model for 100 epochs. All experiments were conducted with PyTorch~\cite{pytorch}.

\subsection{Experiments on VTAB-1K}

\textbf{Datasets.} VTAB-1K~\cite{zhai2019large} is a collection of visual adaptation tasks designed to assess the transferability of pretrained models. It encompasses 19 datasets that can be categorized into three groups: Natural, Specialized, and Structured. Natural datasets consist of images captured by conventional cameras, while Specialized datasets comprise images captured by specialized equipment. Structured datasets evaluate understanding of scene structure, such as object counting or depth estimation. Each dataset includes 800 training images and 200 validation images.

\textbf{Baseline methods.} First we compare our method with full finetuning and linear probing. In full finetuning, all parameters in the model are updated, and linear probing only learns the linear classification head. Second we compared our method with current state-of-the-art PEFT methods, including BitFit~\cite{zaken2021bitfit}, VPT~\cite{jia2022visual}, COMPACTOR \cite{karimi2021compacter}, LoRA~\cite{hu2021lora}, AdaptFormer~\cite{chen2022adaptformer} and FacT~\cite{jie2023fact}. Third, we compared our method with LST~\cite{sung2022lst}, which is also based on side-tuning.

\textbf{Main results.} Results on VTAB-1K are shown in Table \ref{tab:cmp_vtab}. Our LAST demonstrates a notable improvement with 0.9\% higher average accuracy when compared to the previous state-of-the-art method FacT. LAST shows a balanced improvement across three task groups and achieves the top accuracy on 8 out of 19 datasets. Among the three task groups, LAST gains the greatest improvement on the structured datasets, with average accuracy equal to 62.3\% (1.6\% higher than the 60.7\% of FacT).

Apart from the improvement on accuracy, LAST enjoys surprisingly low GPU memory consumption of only 1.33 GB, which is far less than other PEFT methods. Specifically, LAST saves over 75\% and 60\% GPU memory compared to full finetuning and other PEFT methods, respectively. In comparison to the side-tuning method LST (\emph{which is already more memory efficient than other PEFT methods}), LAST only needs roughly half of the GPU memory usage during finetuning, but outperforms LST with a 2.2\% advantage on average accuracy.
Moreover, the number of trainable parameters in LAST is only 28\% of that in LST, indicating that our LAST is much more parameter-efficient than LST, although both are in the side-tuning family. 

% However, LAST indeed requires more trainable parameters than other PEFT methods (\eg, 0.66M vs. COMPACTOR's 0.04M). But, we also want to argue that either the abstract trainable parameter amount (0.66M) or the relative ratio to the pretrained ViT-B ($0.77\%=0.66/85.8$) are both negligible. These small amounts will \emph{not} hinder the deployment of our LAST method in real-world visual adaptation tasks. 

Though LAST requires slightly more trainable parameters than other PEFT methods (\eg, 0.66M vs. COMPACTOR's 0.04M), we argue that in fact the relative ratio of trainable parameters to the pretrained ViT-B ($0.66/85.8=0.77\%$) is still negligible, which does \emph{not} hinder the deployment of LAST in real-world applications.

\subsection{Experiments on FGVC Datasets}

\textbf{Datasets.} Following~\cite{jia2022visual}, we further evaluate our method on five fine-grained visual classification (FGVC) datasets, including CUB-200-2011~\cite{CUB}, NABirds~\cite{van2015building}, Oxford Flowers~\cite{nilsback2008automated}, Stanford Dogs~\cite{khosla2011novel} and Stanford Cars~\cite{gebru2017fine}.

\textbf{Main results.} Results on FGVC datasets are shown in Table \ref{tab:cmp_other}. LAST shows top accuracy on 4 out of 5 datasets and achieves new state-of-the-art among previous PEFT counterparts. The result on FGVC datasets further prove the ability of LAST to finetune on various domains.

\subsection{Experiments on Different Backbones}

\textbf{Settings.} To further explore the scalability of LAST, we compared LAST to full finetuning, linear probing, LoRA and VPT on VTAB-1K with three backbones: ViT-B, ViT-L and ViT-g, which are of different scales. Among these models, ViT-B and ViT-L are pretrained on ImageNet21K~\cite{imagenet}, and ViT-g is pretrained on LVD-142M~\cite{oquab2023dinov2}. In this experiment, we restrict the finetuning process to a single NVIDIA GeForce RTX 3090 GPU with 24 GB memory, which means methods consuming more than 24 GB memory cannot be evaluated.

\textbf{Main results.} From the Results in Table~\ref{tab:cmp_arch}, we observe that LAST consistently outperforms other methods on all three backbones. The accuracy of LAST keeps improving when larger and deeper models are employed as the pretrained backbone $B$.

Most importantly, on the ViT-g backbone with 1167 M parameters, other PEFT methods run out of the memory limit (which is 24 GB). Apart from linear probing, only LST and our LAST can fine-tune under this setting. But LAST outperforms LST with 0.9\% average accuracy gain and significant lower memory consumption. Our LAST only consumes 8.6 GB GPU memory, which is far lower than the 20.4 GB of LST, making finetune ViT-g model possible on a single NVIDIA TITAN Xp GPU (12 GB)!

\begin{table}
    \centering
    \setlength{\tabcolsep}{1.5pt}
    \small
    \begin{tabular}{cc|ccc|c}
        \toprule
        Gap Factor & \#Layer/Block & Natural & Specialized & Structured & Mean \\
        \midrule
            \multirow{2}{*}{$g=1$}
            & $T=1$ & 81.2 & 86.4 & 61.0 & 76.2 \\
            & $T=2$ & 81.1 & 86.5 & 61.9 & 76.5 \\
        \midrule
            \multirow{2}{*}{$g=2$}
            & $T=1$ & 81.0 & 86.2 & 60.7 & 76.0 \\
            & $T=2$ & 81.0 & 86.5 & 62.9 & 76.8 \\
        \midrule
            \multirow{2}{*}{$g=3$}
            & $T=1$ & 81.1 & 86.1 & 59.7 & 75.7 \\
            & $T=2$ & 80.9 & 86.1 & 62.3 & 76.5 \\
        \midrule
            \multirow{2}{*}{$g=4$}
            & $T=1$ & 80.9 & 85.7 & 59.0 & 75.2 \\
            & $T=2$ & 80.9 & 86.0 & 61.7 & 76.2 \\
        \midrule
            \multirow{2}{*}{$g=6$}
            & $T=1$ & 80.9 & 85.1 & 56.4 & 74.1 \\
            & $T=2$ & 80.8 & 85.3 & 58.7 & 74.9 \\
        \bottomrule 
    \end{tabular}
    \caption{Ablation results of network architecture on VTAB-1K with ViT-B/16. The gap factor $g$ decides after how many Transformer blocks a side block is placed. \#Layer/Block denotes the number of LSA modules in every side block.}
    \label{tab:abl_gap}
\end{table}

\begin{table}
    \centering
    \setlength{\tabcolsep}{2.0pt}
    \small
    \begin{tabular}{cc|ccc|c}
        \toprule
        $r_{\rm head}$ & $n_{\rm head}$ & Natural & Specialized & Structured & Mean \\
        \midrule
            \multirow{3}{*}{2}
            & 4 & 80.1 & 85.6 & 60.1 & 75.3 \\
            & 8 & 80.5 & 85.9 & 60.7 & 75.7 \\
            & 16 & 80.8 & 86.0 & 61.9 & 76.3 \\
        \midrule
            \multirow{3}{*}{4}
            & 4 & 80.8 & 85.8 & 61.3 & 76.0 \\
            & 8 & 80.9 & 86.0 & 62.3 & 76.4 \\
            & 16 & 81.2 & 86.2 & 62.3 & 76.6 \\
        \midrule
            \multirow{2}{*}{16}
            & 4 & 81.1 & 86.0 & 61.8 & 76.3 \\
            & 16 & 81.0 & 86.4 & 61.8 & 76.4 \\
        \midrule
            \multirow{2}{*}{64}
            & 1 & 80.9 & 86.2 & 60.6 & 75.9 \\
            & 4 & 80.9 & 86.5 & 63.3 & 76.9 \\
        \midrule
            \multirow{1}{*}{256}
            & 1 & 81.0 & 86.1 & 60.4 & 75.8 \\
        \bottomrule 
    \end{tabular}
    \caption{Ablation on the number of heads and head dimension using VTAB-1K with ViT-B/16. $r_{\rm head}$ is the dimension of attention head. $n_{\rm head}$ is the number of attention heads.}
    \label{tab:abl_multihead}
\end{table}

\subsection{Ablation Studies}

We conduct ablation studies on VTAB-1K to investigate the impact of different design choices in LAST. 

\noindent\textbf{Varying gap factor $g$ and stack factor $T$.} In our LAST, the gap factor $g$ affects the number of side blocks, and the stack factor $T$ decides the number of layers per side block contains. A smaller $g$ and a larger $T$ means more parameters and greater memory usage. We empirically study how many side blocks and how many layers per block is enough for adaptation. The result in Table~\ref{tab:abl_gap} shows that more parameters generally result in better accuracy. For $T=1$, using $g=1$ leads to the best average accuracy of 76.2\%. For $T=2$, the peak accuracy is observed when $g=2$, achieving an average accuracy of 76.8\%.

\noindent \textbf{The impact of head dimension $r_{head}$ and head number $n_{head}$.} To verify the low-rank hypothesis of residual side-tuning, we experiment on different head dimension $r_{head}$ and number of heads $n_{head}$ in side block attention with a fixed $g=3$, $T=2$. The results in Table~\ref{tab:abl_multihead} reveal several important findings. First, when the head dimension is fixed, we can almost always gain some improvement by increasing the number of heads. But, if we fix the number of heads, increasing the head dimension will not necessarily result in higher accuracy. For example, using $r_{\rm head}=256$ and $n_{\rm head}=1$ only leads to 75.8\% average accuracy, which is significantly lower than the result of $r_{\rm head}=4$ and $n_{\rm head}=8$. On most downstream tasks, setting head dimension to 4 is already enough.

\begin{figure}
    \centering
    \includegraphics[width=0.7\linewidth]{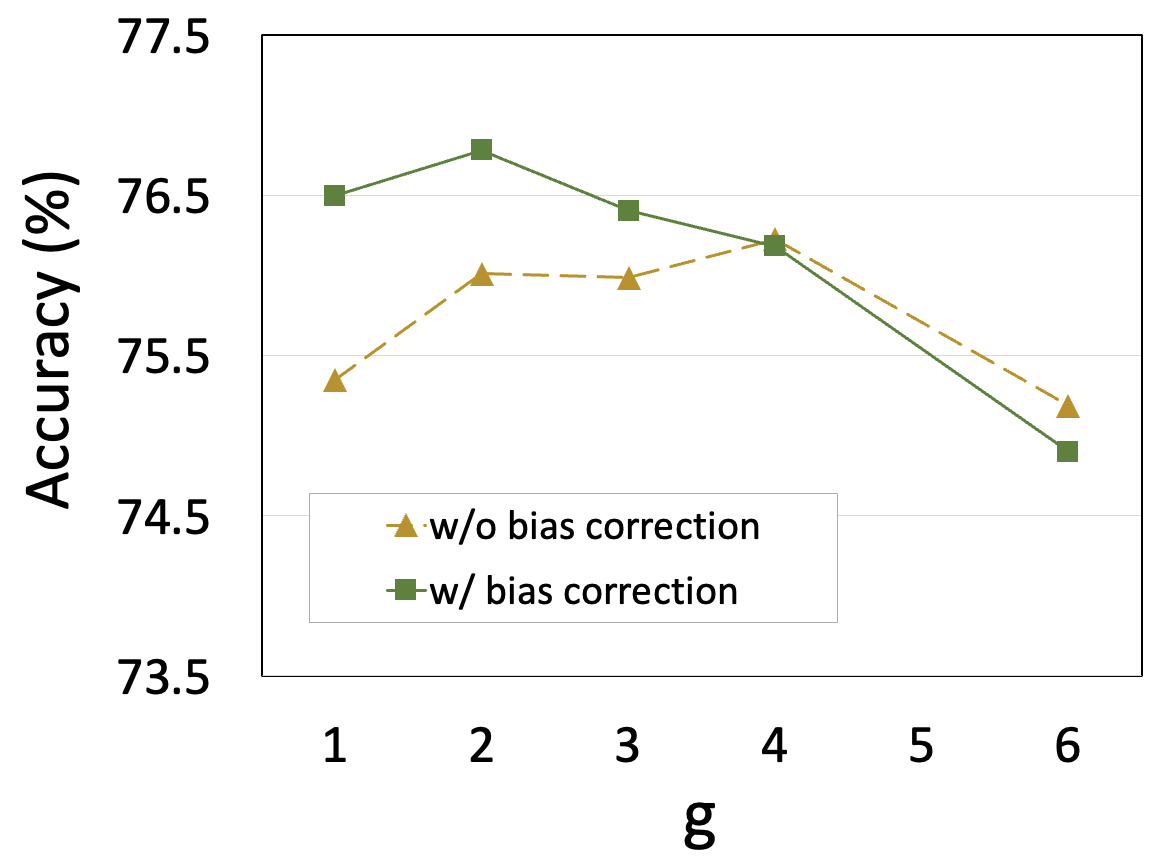}
    \caption{Effect of the bias correction in the LSA module. Y-axis represents the group-wise average accuracy on VTAB-1K.}
    \label{fig:subtract}
\end{figure}

\noindent\textbf{The impact of LSA bias correction.} As discussed in Section~\ref{ssec:subtract}, we have showed that it is important to correct the bias in our LSA module. In Figure~\ref{fig:subtract}, we test the effectiveness of bias correction. We find that when the gap factor $g$ is small (such that more intermediate features are added), bias correction can improve the performance by a large margin (\eg, 1.2\% improvement when $g=1$). When $g$ grows larger ($g=4,6$), leaving the intermediate feature not subtracted will not harm overall performance.

\section{Conclusions and Future Work}

We proposed an effective finetuning method LAST, which possesses merits of both PEFT and side-tuning. It significantly reduces memory usage and time consumption during finetuning and achieves superior accuracy over previous state-of-the-art PEFT methods. We showed that a side network based on low-rank attention can well learns task-specific knowledge even if the backbone features are frozen. We believe this finding can inspire more efficient finetuning methods. We also proved that intermediate features added to the side-network should be corrected as to recover backbone representation. Given its small GPU memory footprint, fast training speed, parallel training capability, parameter-efficiency and high accuracy, LAST is an excellent choice for visual adaptation.

One limitation is that it is not convenient to transfer LAST to other backbone networks. It is worthwhile extending LAST to other models (\eg, ResNet, DenseNet) and other visual adaptation tasks (\eg, object detection and image generation). It is also promising to apply LAST on large language models, although our GPU resources are limited to conduct those experiments.

\newpage
\setcounter{page}{1}
\setcounter{figure}{3}
\setcounter{table}{5}
\setcounter{section}{0}
\maketitlesupplementary

\section{Ablations on FFN in LAST}

To further explore the effectiveness of using only low-rank attention in LAST, we conduct ablation studies on using feed-forward network (FFN) in LAST, following default settings on VTAB-1K with ViT-B backbone. Like stardard ViTs, we use GELU non-linearity in FFN. Results are shown in Table~\ref{tab:ffn}. We can observe that replacing low-rank attention with FFN will cause dramatic drop of average accuracy no matter what hidden dimension we choose. On the other hand, adding an FFN after each attention module will not bring about noticeable improvement, the average accuracy may even drop slightly when $h=256$. Meanwhile, the large FFN poses extra parameters and memory consumption to the side network, which can not be neglected.

These results empirically corroborate the design of using only low-rank attention in side network.

\begin{table}[H]
    \centering
    \setlength{\tabcolsep}{4pt}
    \small
    \begin{tabular}{ccc|ccc}
        \toprule
        Attention & FFN & $h$ & Mean Acc & Mem (GB) & Params (M) \\
        \midrule
         & & & 57.6 & 0.57 & 0 \\
         & $\checkmark$ & 64 & 72.1 & 1.19 & 1.32 \\
         & $\checkmark$ & 256 & 72.9 & 1.39 & 5.29 \\
         \midrule
        $\checkmark$ & & & 76.5 & 1.33 & 0.66 \\
        $\checkmark$ & $\checkmark$ & 64 & 76.6 & 1.90 & 1.98 \\
        $\checkmark$ & $\checkmark$ & 256 & 76.3 & 2.08 & 5.95 \\
        \bottomrule 
    \end{tabular}
    \caption{Ablation results on whether using FFN in side network, $h$ decides the hidden dimension in FFN. ``Mean acc'' is group-wise average accuracy over three task groups of VTAB-1K. Using neither attention nor FFN in side network denotes linear probing. And using only attention denotes default LAST.}
    \label{tab:ffn}
\end{table}

\section{Effect of increasing stack factor $T$}

\begin{figure}
    \centering
    \includegraphics[width=0.9\linewidth]{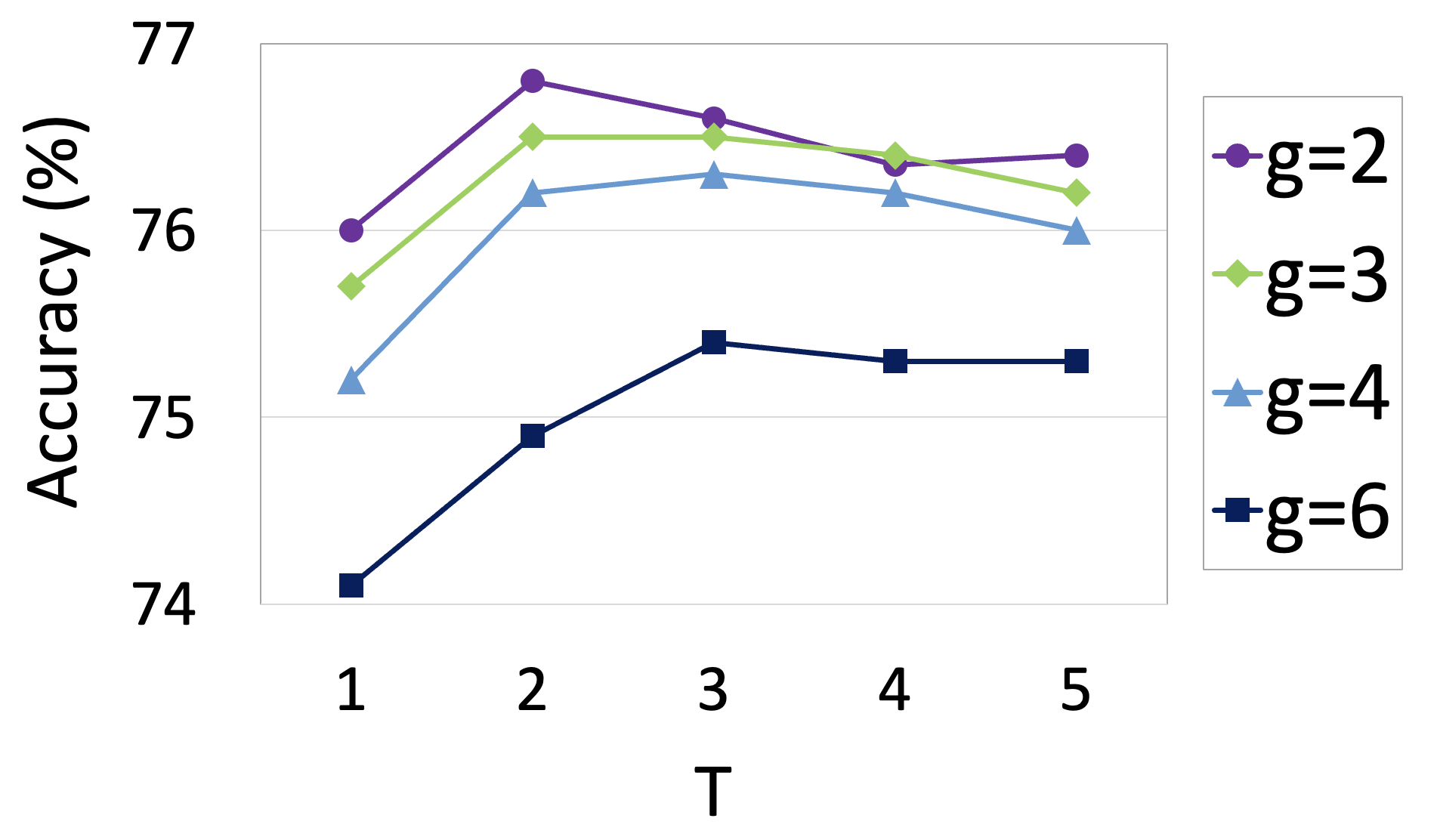}
    \caption{Effect of varying $T$ in LAST. Y-axis represents the group-wise average accuracy on VTAB-1K.}
    \label{fig:abl_t}
\end{figure}

In this part, we further test the performance of LAST when a larger stack factor $T$ is adopted (\ie, $T>2$). We test different choices of $T$ from 1 to 5 with the default setting on VTAB-1K using ViT-B as backbone. As shown in Figure~\ref{fig:abl_t}, the overall accuracy gets higher when we increase $T$ from 1 to 2. But in most cases (\ie, $g=2,3,4$), overall accuracy do not benefit from a larger $T$, and may even drop when $T$ keeps increasing. The only exception is when $g=6$, $T=3$ leads to the best average accuracy of $75.4\%$, but this is still far below results from taking a smaller gap factor $g$. Therefore, selecting $T=2$ is enough for LAST to work properly in most cases.

\section{Training speed and GPU memory usage}

We list the training speed and GPU memory usage during finetuning of multiple methods in detail. We use \textit{torch.cuda.max\_memory\_allocated()} function to record peck memory usage during finetuning. Results are given in Table~\ref{tab:speed_mem}. LAST enjoys significantly faster training speed and smaller GPU memory usage compared to other finetuning methods.

\begin{table}
    \centering
    \setlength{\tabcolsep}{7pt}
    \small
    \begin{tabular}{l|cc}
        \toprule
        Method & Training speed (ms/batch) & GPU mem (GB) \\
        \midrule
        Full finetuning & 582 & 6.09 \\
        Linear probing & 205 & 0.57 \\
        \midrule
        LoRA~\cite{hu2021lora_s} & 525 & 3.40 \\
        FacT~\cite{jie2023fact_s} & 398 & 4.81 \\
        BitFit~\cite{zaken2021bitfit_s} & 438 & 3.80 \\
        VPT-Deep~\cite{jia2022visual_s} & 603 & 5.63 \\
        LST~\cite{sung2022lst_s} & 364 & 2.65 \\
        \midrule
        LAST (\textbf{ours}) & \textbf{281} & \textbf{1.33} \\
        \bottomrule 
    \end{tabular}
    \caption{Traning speed and GPU memory usage comparison between different finetuning methods. Experiments are conducted on ViT-B with batch size 32, image size $224\times224$ and without mixed precision training on an NVIDIA TITAN-Xp GPU.}
    \label{tab:speed_mem}
\end{table}

\section{Visualization on VTAB datasets}

\begin{figure*}
    \centering
    \includegraphics[width=\linewidth]{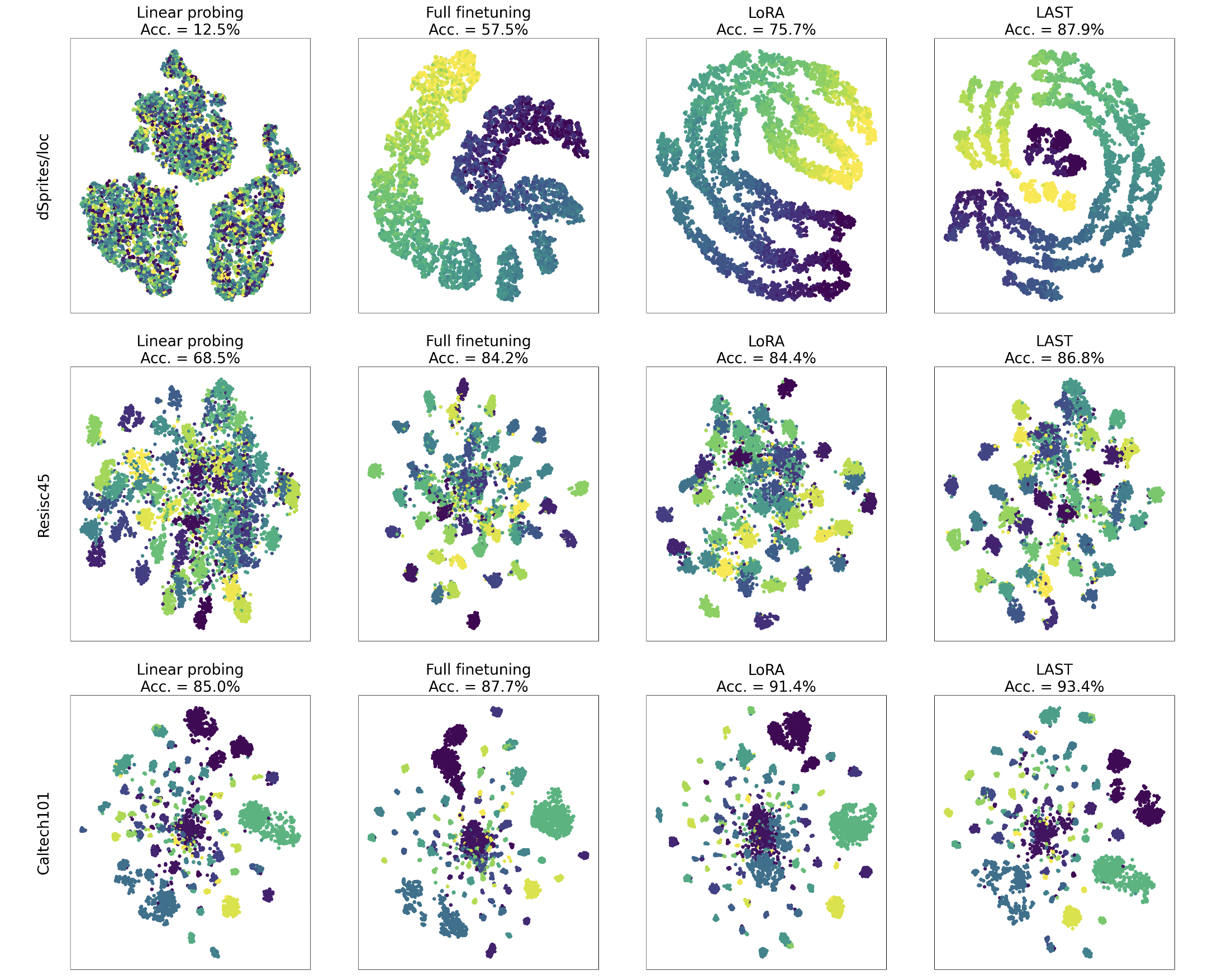}
    \caption{t-SNE Visualization of feataure distribution from four different finetuning methods: linear probing, full finetuning, LoRA and our LAST. ``dSprites/loc'' dataset comes from ``Structured'' group of VTAB-1K, ``Resisc45'' from ``Specialized'' group and ``Caltech101'' from ``Natural'' group. Dots of different colors belong to different categories. (Best viewed in color.)}
    \label{fig:tsne}
\end{figure*}

We also visualize the feature distribution from different finetuning methods via t-SNE~\cite{maaten2008tsne_s} on VTAB datasets. We choose three downstream tasks from three task groups in VTAB-1K. Results are shown in Figure~\ref{fig:tsne}. LAST achieves better clustering results compared to linear probing, full finetuning and another PEFT method LoRA.

\begin{thebibliography}{34}
\small
\providecommand{\natexlab}[1]{#1}
\providecommand{\url}[1]{\texttt{#1}}
\expandafter\ifx\csname urlstyle\endcsname\relax
  \providecommand{\doi}[1]{doi: #1}\else
  \providecommand{\doi}{doi: \begingroup \urlstyle{rm}\Url}\fi

\bibitem[Ba et~al.(2016)Ba, Kiros, and Hinton]{ba2016layer}
Jimmy~Lei Ba, Jamie~Ryan Kiros, and Geoffrey~E Hinton.
\newblock Layer normalization.
\newblock \emph{arXiv preprint arXiv:1607.06450}, 2016.

\bibitem[Baevski and Auli(2019)]{baevski2018adaptive}
Alexei Baevski and Michael Auli.
\newblock Adaptive input representations for neural language modeling.
\newblock In \emph{International Conference on Learning Representations}, pages 1--11, 2019.

\bibitem[Chen et~al.(2022)Chen, Ge, Tong, Wang, Song, Wang, and Luo]{chen2022adaptformer}
Shoufa Chen, Chongjian Ge, Zhan Tong, Jiangliu Wang, Yibing Song, Jue Wang, and Ping Luo.
\newblock {Adaptformer: Adapting vision transformers for scalable visual recognition}.
\newblock In \emph{Advances in Neural Information Processing Systems}, pages 16664--16678, 2022.

\bibitem[Dosovitskiy et~al.(2021)Dosovitskiy, Beyer, Kolesnikov, Weissenborn, Zhai, Unterthiner, Dehghani, Minderer, Heigold, Gelly, Uszkoreit, and Houlsby]{dosovitskiy2020image}
Alexey Dosovitskiy, Lucas Beyer, Alexander Kolesnikov, Dirk Weissenborn, Xiaohua Zhai, Thomas Unterthiner, Mostafa Dehghani, Matthias Minderer, Georg Heigold, Sylvain Gelly, Jakob Uszkoreit, and Neil Houlsby.
\newblock An image is worth 16x16 words: {Transformers} for image recognition at scale.
\newblock In \emph{International Conference on Learning Representations}, pages 1--21, 2021.

\bibitem[Gebru et~al.(2017)Gebru, Krause, Wang, Chen, Deng, and Fei-Fei]{gebru2017fine}
Timnit Gebru, Jonathan Krause, Yilun Wang, Duyun Chen, Jia Deng, and Li Fei-Fei.
\newblock Fine-grained car detection for visual census estimation.
\newblock In \emph{Proceedings of the AAAI Conference on Artificial Intelligence}, 2017.

\bibitem[He et~al.(2016)He, Zhang, Ren, and Sun]{he2016residual}
Kaiming He, Xiangyu Zhang, Shaoqing Ren, and Jian Sun.
\newblock Deep residual learning for image recognition.
\newblock In \emph{Proceedings of the IEEE Conference on Computer Vision and Pattern Recognition}, pages 770--778, 2016.

\bibitem[He et~al.(2022)He, Chen, Xie, Li, Doll\'ar, and Girshick]{he2022masked}
Kaiming He, Xinlei Chen, Saining Xie, Yanghao Li, Piotr Doll\'ar, and Ross Girshick.
\newblock Masked autoencoders are scalable vision learners.
\newblock In \emph{Proceedings of the IEEE/CVF Conference on Computer Vision and Pattern Recognition}, pages 16000--16009, 2022.

\bibitem[He et~al.(2023)He, Li, Zhang, Yang, and Wang]{he2023parameter}
Xuehai He, Chunyuan Li, Pengchuan Zhang, Jianwei Yang, and Xin~Eric Wang.
\newblock Parameter-efficient model adaptation for vision transformers.
\newblock In \emph{Proceedings of the AAAI Conference on Artificial Intelligence}, pages 817--825, 2023.

\bibitem[Hinton et~al.(2015)Hinton, Vinyals, and Dean]{hinton2015distilling}
Geoffrey Hinton, Oriol Vinyals, and Jeff Dean.
\newblock Distilling the knowledge in a neural network.
\newblock \emph{arXiv preprint arXiv:1503.02531}, 2015.

\bibitem[Houlsby et~al.(2019)Houlsby, Giurgiu, Jastrzebski, Morrone, De~Laroussilhe, Gesmundo, Attariyan, and Gelly]{houlsby2019parameter}
Neil Houlsby, Andrei Giurgiu, Stanislaw Jastrzebski, Bruna Morrone, Quentin De~Laroussilhe, Andrea Gesmundo, Mona Attariyan, and Sylvain Gelly.
\newblock Parameter-efficient transfer learning for {NLP}.
\newblock In \emph{International Conference on Machine Learning}, pages 2790--2799, 2019.

\bibitem[Hu et~al.(2022)Hu, Shen, Wallis, Allen-Zhu, Li, Wang, Wang, and Chen]{hu2021lora}
Edward~J Hu, Yelong Shen, Phillip Wallis, Zeyuan Allen-Zhu, Yuanzhi Li, Shean Wang, Lu Wang, and Weizhu Chen.
\newblock {LoRA}: {L}ow-rank adaptation of large language models.
\newblock In \emph{International Conference on Learning Representation}, pages 1--13, 2022.

\bibitem[Jia et~al.(2022)Jia, Tang, Chen, Cardie, Belongie, Hariharan, and Lim]{jia2022visual}
Menglin Jia, Luming Tang, Bor-Chun Chen, Claire Cardie, Serge Belongie, Bharath Hariharan, and Ser-Nam Lim.
\newblock {Visual Prompt Tuning}.
\newblock In \emph{European Conference on Computer Vision}, pages 709--727. Springer, 2022.

\bibitem[Jie and Deng(2023)]{jie2023fact}
Shibo Jie and Zhi-Hong Deng.
\newblock {FacT}: {F}actor-tuning for lightweight adaptation on vision transformer.
\newblock In \emph{Proceedings of the AAAI Conference on Artificial Intelligence}, pages 1060--1068, 2023.

\bibitem[Karimi~Mahabadi et~al.(2021)Karimi~Mahabadi, Henderson, and Ruder]{karimi2021compacter}
Rabeeh Karimi~Mahabadi, James Henderson, and Sebastian Ruder.
\newblock {Compacter: Efficient} low-rank hypercomplex adapter layers.
\newblock pages 1022--1035, 2021.

\bibitem[Khosla et~al.(2011)Khosla, Jayadevaprakash, Yao, and Li]{khosla2011novel}
Aditya Khosla, Nityananda Jayadevaprakash, Bangpeng Yao, and Fei-Fei Li.
\newblock Novel dataset for fine-grained image categorization.
\newblock In \emph{First Workshop on Fine-Grained Visual Categorization, IEEE Conference on Computer Vision and Pattern Recognition}, 2011.

\bibitem[Kingma and Ba(2015)]{kingma2014adam}
Diederik~P Kingma and Jimmy Ba.
\newblock {Adam: A method for stochastic optimization}.
\newblock In \emph{International Conference on Learning Representations}, pages 1--11, 2015.

\bibitem[Li and Liang(2021)]{li2021prefix}
Xiang~Lisa Li and Percy Liang.
\newblock {Prefix-Tuning: Optimizing} continuous prompts for generation.
\newblock In \emph{Proceedings of the 59th Annual Meeting of the Association for Computational Linguistics}, pages 4582--4597, 2021.

\bibitem[Lian et~al.(2022)Lian, Zhou, Feng, and Wang]{lian2022scaling}
Dongze Lian, Daquan Zhou, Jiashi Feng, and Xinchao Wang.
\newblock {Scaling \&amp; Shifting Your Features: A new baseline for efficient model tuning}.
\newblock In \emph{Advances in Neural Information Processing Systems}, pages 109--123, 2022.

\bibitem[Nilsback and Zisserman(2008)]{nilsback2008automated}
Maria-Elena Nilsback and Andrew Zisserman.
\newblock Automated flower classification over a large number of classes.
\newblock In \emph{Indian Conference on Computer Vision, Graphics \& Image Processing}, pages 722--729, 2008.

\bibitem[Oquab et~al.(2023)Oquab, Darcet, Moutakanni, Vo, Szafraniec, Khalidov, Fernandez, Haziza, Massa, El-Nouby, et~al.]{oquab2023dinov2}
Maxime Oquab, Timoth{\'e}e Darcet, Th{\'e}o Moutakanni, Huy Vo, Marc Szafraniec, Vasil Khalidov, Pierre Fernandez, Daniel Haziza, Francisco Massa, Alaaeldin El-Nouby, et~al.
\newblock Dinov2: Learning robust visual features without supervision.
\newblock \emph{arXiv preprint arXiv:2304.07193}, 2023.

\bibitem[Paszke et~al.(2019)Paszke, Gross, Massa, Lerer, Bradbury, Chanan, Killeen, Lin, Gimelshein, Antiga, Desmaison, Kopf, Yang, DeVito, Raison, Tejani, Chilamkurthy, Steiner, Fang, Bai, and Chintala]{pytorch}
Adam Paszke, Sam Gross, Francisco Massa, Adam Lerer, James Bradbury, Gregory Chanan, Trevor Killeen, Zeming Lin, Natalia Gimelshein, Luca Antiga, Alban Desmaison, Andreas Kopf, Edward Yang, Zachary DeVito, Martin Raison, Alykhan Tejani, Sasank Chilamkurthy, Benoit Steiner, Lu Fang, Junjie Bai, and Soumith Chintala.
\newblock {PyTorch}: {An} imperative style, high-performance deep learning library.
\newblock In \emph{Advances in Neural Information Processing Systems}, pages 8026--8037, 2019.

\bibitem[Radford et~al.(2021)Radford, Kim, Hallacy, Ramesh, Goh, Agarwal, Sastry, Askell, Mishkin, Clark, et~al.]{radford2021learning}
Alec Radford, Jong~Wook Kim, Chris Hallacy, Aditya Ramesh, Gabriel Goh, Sandhini Agarwal, Girish Sastry, Amanda Askell, Pamela Mishkin, Jack Clark, et~al.
\newblock Learning transferable visual models from natural language supervision.
\newblock In \emph{Proceedings of the International Conference on Machine Learning}, pages 8748--8763, 2021.

\bibitem[Raffel et~al.(2020)Raffel, Shazeer, Roberts, Lee, Narang, Matena, Zhou, Li, and Liu]{raffel2020exploring}
Colin Raffel, Noam Shazeer, Adam Roberts, Katherine Lee, Sharan Narang, Michael Matena, Yanqi Zhou, Wei Li, and Peter~J Liu.
\newblock Exploring the limits of transfer learning with a unified text-to-text transformer.
\newblock \emph{The Journal of Machine Learning Research}, 21\penalty0 (140):\penalty0 1--67, 2020.

\bibitem[Russakovsky et~al.(2015)Russakovsky, Deng, Su, Krause, Satheesh, Ma, Huang, Karpathy, Khosla, Bernstein, Berg, and Fei-Fei]{imagenet}
Olga Russakovsky, Jia Deng, Hao Su, Jonathan Krause, Sanjeev Satheesh, Sean Ma, Zhiheng Huang, Andrej Karpathy, Aditya Khosla, Michael Bernstein, Alexander~C. Berg, and Li Fei-Fei.
\newblock {ImageNet} large scale visual recognition challenge.
\newblock \emph{IJCV}, 115\penalty0 (3):\penalty0 211--252, 2015.

\bibitem[Sung et~al.(2022{\natexlab{a}})Sung, Cho, and Bansal]{sung2022lst}
Yi-Lin Sung, Jaemin Cho, and Mohit Bansal.
\newblock {LST: Ladder Side-Tuning} for parameter and memory efficient transfer learning.
\newblock In \emph{Advances in Neural Information Processing Systems}, pages 12991--13005, 2022{\natexlab{a}}.

\bibitem[Sung et~al.(2022{\natexlab{b}})Sung, Cho, and Bansal]{sung2022vl}
Yi-Lin Sung, Jaemin Cho, and Mohit Bansal.
\newblock {VL-Adapter}: {Parameter}-efficient transfer learning for vision-and-language tasks.
\newblock In \emph{Proceedings of the IEEE/CVF Conference on Computer Vision and Pattern Recognition}, pages 5227--5237, 2022{\natexlab{b}}.

\bibitem[Van~Horn et~al.(2015)Van~Horn, Branson, Farrell, Haber, Barry, Ipeirotis, Perona, and Belongie]{van2015building}
Grant Van~Horn, Steve Branson, Ryan Farrell, Scott Haber, Jessie Barry, Panos Ipeirotis, Pietro Perona, and Serge Belongie.
\newblock Building a bird recognition app and large scale dataset with citizen scientists: The fine print in fine-grained dataset collection.
\newblock In \emph{IEEE Conference on Computer Vision and Pattern Recognition}, pages 595--604, 2015.

\bibitem[Vaswani et~al.(2017)Vaswani, Shazeer, Parmar, Uszkoreit, Jones, Gomez, Kaiser, and Polosukhin]{vaswani2017attention}
Ashish Vaswani, Noam Shazeer, Niki Parmar, Jakob Uszkoreit, Llion Jones, Aidan~N Gomez, {\L}ukasz Kaiser, and Illia Polosukhin.
\newblock Attention is all you need.
\newblock In \emph{Advances in Neural Information Processing Systems}, pages 6000--6010, 2017.

\bibitem[Wah et~al.(2011)Wah, Branson, Welinder, Perona, and Belongie]{CUB}
Catherine Wah, Steve Branson, Peter Welinder, Pietro Perona, and Serge Belongie.
\newblock {The Caltech-UCSD Birds-200-2011 Dataset}.
\newblock Technical Report CNS-TR-2011-001, California Institute of Technology, 2011.

\bibitem[Wang et~al.(2019)Wang, Li, Xiao, Zhu, Li, Wong, and Chao]{wang2019learning}
Qiang Wang, Bei Li, Tong Xiao, Jingbo Zhu, Changliang Li, Derek~F Wong, and Lidia~S Chao.
\newblock Learning deep transformer models for machine translation.
\newblock In \emph{Proceedings of the 57th Annual Meeting of the Association for Computational Linguistics}, pages 1810--1822, 2019.

\bibitem[Yin et~al.(2023)Yin, Yang, Wang, Yu, Wei, and Sun]{yin2023one}
Dongshuo Yin, Yiran Yang, Zhechao Wang, Hongfeng Yu, Kaiwen Wei, and Xian Sun.
\newblock 1\% vs 100\%: Parameter-efficient low rank adapter for dense predictions.
\newblock In \emph{Proceedings of the IEEE/CVF Conference on Computer Vision and Pattern Recognition}, pages 20116--20126, 2023.

\bibitem[Zaken et~al.(2022)Zaken, Ravfogel, and Goldberg]{zaken2021bitfit}
Elad~Ben Zaken, Shauli Ravfogel, and Yoav Goldberg.
\newblock {B}it{F}it: {S}imple parameter-efficient fine-tuning for transformer-based masked language-models.
\newblock In \emph{Proceedings of the 60th Annual Meeting of the Association for Computational Linguistics}, pages 1--9, 2022.

\bibitem[Zhai et~al.(2019)Zhai, Puigcerver, Kolesnikov, Ruyssen, Riquelme, Lucic, Djolonga, Pinto, Neumann, Dosovitskiy, et~al.]{zhai2019large}
Xiaohua Zhai, Joan Puigcerver, Alexander Kolesnikov, Pierre Ruyssen, Carlos Riquelme, Mario Lucic, Josip Djolonga, Andre~Susano Pinto, Maxim Neumann, Alexey Dosovitskiy, et~al.
\newblock A large-scale study of representation learning with the visual task adaptation benchmark.
\newblock \emph{arXiv preprint arXiv:1910.04867}, 2019.

\bibitem[Zhang et~al.(2020)Zhang, Sax, Zamir, Guibas, and Malik]{zhang2020side}
Jeffrey~O Zhang, Alexander Sax, Amir Zamir, Leonidas Guibas, and Jitendra Malik.
\newblock {Side-Tuning: a baseline for network adaptation via additive side networks}.
\newblock In \emph{European Conference on Computer Vision}, pages 698--714. Springer, 2020.

\end{thebibliography}

\begin{thebibliography}{6}
\small
\providecommand{\natexlab}[1]{#1}
\providecommand{\url}[1]{\texttt{#1}}
\expandafter\ifx\csname urlstyle\endcsname\relax
  \providecommand{\doi}[1]{doi: #1}\else
  \providecommand{\doi}{doi: \begingroup \urlstyle{rm}\Url}\fi

\bibitem[Hu et~al.(2022)Hu, Shen, Wallis, Allen-Zhu, Li, Wang, Wang, and Chen]{hu2021lora_s}
Edward~J Hu, Yelong Shen, Phillip Wallis, Zeyuan Allen-Zhu, Yuanzhi Li, Shean Wang, Lu Wang, and Weizhu Chen.
\newblock {LoRA}: {L}ow-rank adaptation of large language models.
\newblock In \emph{International Conference on Learning Representation}, pages 1--13, 2022.

\bibitem[Jia et~al.(2022)Jia, Tang, Chen, Cardie, Belongie, Hariharan, and Lim]{jia2022visual_s}
Menglin Jia, Luming Tang, Bor-Chun Chen, Claire Cardie, Serge Belongie, Bharath Hariharan, and Ser-Nam Lim.
\newblock {Visual Prompt Tuning}.
\newblock In \emph{European Conference on Computer Vision}, pages 709--727. Springer, 2022.

\bibitem[Jie and Deng(2023)]{jie2023fact_s}
Shibo Jie and Zhi-Hong Deng.
\newblock {FacT}: {F}actor-tuning for lightweight adaptation on vision transformer.
\newblock In \emph{Proceedings of the AAAI Conference on Artificial Intelligence}, pages 1060--1068, 2023.

\bibitem[Sung et~al.(2022)Sung, Cho, and Bansal]{sung2022lst_s}
Yi-Lin Sung, Jaemin Cho, and Mohit Bansal.
\newblock {LST: Ladder Side-Tuning} for parameter and memory efficient transfer learning.
\newblock In \emph{Advances in Neural Information Processing Systems}, pages 12991--13005, 2022.

\bibitem[van~der Maaten and Hinton(2008)]{maaten2008tsne_s}
Laurens van~der Maaten and Geoffrey Hinton.
\newblock Visualizing data using t-sne.
\newblock \emph{Journal of Machine Learning Research}, 9\penalty0 (86):\penalty0 2579--2605, 2008.

\bibitem[Zaken et~al.(2022)Zaken, Ravfogel, and Goldberg]{zaken2021bitfit_s}
Elad~Ben Zaken, Shauli Ravfogel, and Yoav Goldberg.
\newblock {B}it{F}it: {S}imple parameter-efficient fine-tuning for transformer-based masked language-models.
\newblock In \emph{Proceedings of the 60th Annual Meeting of the Association for Computational Linguistics}, pages 1--9, 2022.

\end{thebibliography}
\end{document}